\documentclass[
]{ceurart}

\usepackage{listings}
\usepackage{inputenc}
\begin{document}

\copyrightyear{2022}
\copyrightclause{Copyright for this paper by its authors.
  Use permitted under Creative Commons License Attribution 4.0
  International (CC BY 4.0).}

\conference{MediaEval'22: Multimedia Evaluation Workshop,
  January 13--15, 2023, Bergen, Norway and Online}
\title{Sport Task: Fine Grained Action Detection and Classification of Table Tennis Strokes from Videos for MediaEval 2022}

\author[1]{Pierre-Etienne Martin}[%
    orcid=0000-0002-9593-4580,
    email=mediaeval.sport.task@diff.u-bordeaux.fr,
    url=www.eva.mpg.de/comparative-cultural-psychology/staff/pierre-etienne-martin,
]
\author[2]{Jordan Calandre}
\author[3]{Boris Mansencal}
\author[3]{Jenny Benois-Pineau}
\author[2]{Renaud P\'eteri}
\author[2]{Laurent Mascarilla}
\author[4]{Julien Morlier}

\address[1]{CCP Department, Max Planck Institute for Evolutionary Anthropology, D-04103 Leipzig, Germany}
\address[2]{MIA, La Rochelle University, La Rochelle, France}
\address[3]{Univ. Bordeaux, CNRS,  Bordeaux INP, LaBRI, Talence, France}
\address[4]{IMS, University of Bordeaux, Talence, France}


\begin{abstract}
Sports video analysis is a widespread research topic. Its applications are very diverse, like events detection during a match, video summary, or fine-grained movement analysis of athletes. As part of the MediaEval 2022 benchmarking initiative, this task aims at detecting and classifying subtle movements from sport videos. We focus on recordings of table tennis matches.
Conducted since 2019, this task provides a classification challenge from untrimmed videos recorded under natural conditions with known temporal boundaries for each stroke.
Since 2021, the task also provides a stroke detection challenge from un-annotated, untrimmed videos.
This year, the training, validation, and test sets are enhanced to ensure that all strokes are represented in each dataset. The dataset is now similar to the one used in~\cite{PeICPR:2020,PeThesis2020}.
This research is intended to build tools for coaches and athletes who want to further evaluate their sport performances. 
\end{abstract}

\maketitle

\section{Introduction}
\label{sec:intro}
Action detection and classification is one of the main challenges in computer vision~\cite{PeChapSpringer:2021}. Throughout the last few years, datasets focusing on action classification have grown tremendously, along with their complexity~\cite{PeThesis2020}.
There has been a significant number of studies devoted to the analysis of sports gestures using motion-capture systems. Nonetheless, sensors and markers attached to the body have the inherent tendency to interfere with the natural behaviour of the athletes. Therefore, this concern motivates the development of non-invasive methods using video recording from a camera.
\par
The sports video classification project was initiated by the Sports Faculty of the University of Bordeaux (STAPS), the computer science laboratory LaBRI, and the MIA laboratory of the University of La Rochelle \footnote{This research is supported by the New Aquitania Region's CRISP project - ComputeR vIsion for Sports Performance and the MIRES Federation.}. This project intends to develop  artificial intelligence and multimedia indexing methods for table tennis stroke recognition. The ultimate goal is to evaluate the performance of individual athletes, especially students, and thereby develop optimal training strategies. For this purpose, we have recorded a video corpus \texttt{TTStroke-21} with volunteer athletes.
\par
Datasets like UCF-101~\cite{Dataset:UCF101:2012}, HMDB~\cite{Dataset:HMDB:2011,Dataset:JHMDB:2013}, AVA~\cite{Dataset:AVA:2018} and Kinetics~\cite{Dataset:Kinetics:2017,Dataset:Kinetics700:2020,Dataset:AVA_Kinetics:2020} are used in the action recognition field, with an increasing number of video samples and number of classes covered over the years. There are very few datasets available that have a focus on fine-grained classifications in sports, such as FineGym~\cite{Dataset:Gym:2020} and TTStroke21~\cite{PeMTAP:2020}.
\par
To address the increasing complexity of datasets, some classification methods exploit temporal information as much as possible. For example, \cite{LiuH:2019}~learns spatio-temporal dependencies from videos using only RGB data. Alternatively, some methods integrate other modalities extracted from videos, e.g., optical flow~\cite{NN:I3DCarreira:2017,NN:Laptev:2018, PeICIP:2019}. Moreover, in the \texttt{TTStroke-21} dataset, stroke classification is challenging, because movements between two strokes share strong visual similarities.
\par
The following sections present the Sport task and its substasks for this year, along with the dataset and the specific terms of use when downloading the dataset. Complementary information on the task may be found on the dedicated page from the MediaEval website\footnote{\url{https://multimediaeval.github.io/editions/2022/tasks/sportsvideo/}}.

\section{Task description}
\label{sec:task}

The Sport task is based on the \texttt{TTStroke-21} database~\cite{PeCBMI:2018, PeMTAP:2020}. This database is a corpus of table tennis recordings with players performing in natural conditions. The dataset delivered through this task focuses on videos acquired with GoPro cameras at a recording speed of 120 frames per second and annotated by professional players. This task offers researchers an opportunity to solve a fine-grained classification problem with videos and annotations of high quality in the sports domain. Compared to the Sport task from MediaEval 2021's edition~\cite{DBLP:conf/mediaeval/MartinCMBPMM21}, the dataset has been enriched and the data organization and distribution differ. The task has two subtasks: stroke classification from trimmed videos and stroke detection from untrimmed videos. Each subtask has its own dataset.
\par
Researchers can participate in one or both subtasks and submit up to five runs for each subtask. The participants must fill in the provided XML files dedicated to the test set of the subtask for each run. The content of the XML file varies according to the subtask. The runs have to be submitted in an archive (zip file), with each run in a different directory for each subtask. Participants should also submit a working notes paper, which describes their method and indicates if any external data, such as other datasets or pre-trained networks, was used to compute their runs. The use of pre-trained models on the Sport task dataset \texttt{TTStroke-21} of the previous years is however forbidden.  The task is considered fully automatic: once the videos are provided to the system, results should be produced without any human intervention. Participants are encouraged to release their code publicly with their submission. This year, similarly to the 2021 edition, a baseline for both subtasks is shared publicly\footnote{\url{https://github.com/ccp-eva/SportTaskME22}}~\cite{mediaeval/Martin/2022/baseline}.

\subsection{Subtask 1 - Stroke Classification}

 For this subtask, the participants are required to classify a set of trimmed videos containing only one table tennis stroke, or possibly no stroke at all. There are 20 possible stroke classes and an additional non-stroke class. For this purpose, two annotated sets are provided: a training and a validation set with respectively 807 and 230 trimmed videos. A non-annotated test set comprising 118 trimmed videos has to be classified. The trimmed videos in the different sets may have been retrieved from the same untrimmed videos but at different moments in time without overlapping.
 \par
 Specifically, the participants are invited to fill an XML file and replace the default label ``\texttt{Unknown}'' with the stroke class assigned by the participants' method. All submissions will be evaluated in terms of global accuracy for ranking, and detailed with per-class accuracy.
\par
In last year edition, the best global accuracy reached 74.2\%~\cite{DBLP:conf/mediaeval/QianYL021} using SWIN-Transformers, followed closely by ResNet-50 models (68.8\%) beating by a large margin last year baseline (20.4\%)~\cite{DBLP:conf/mediaeval/Martin21}. Methods effectiveness seem to be linked to the model architecture, but also the taking into account of both stroke class similarity and class imbalance during training. This year, the task uses the same split of the \texttt{TTStroke-21} database as in \cite{PeICPR:2020,PeThesis2020} allowing better comparison with previous works outside the MediaEval benchmark scope.

\subsection{Subtask 2 - Stroke Detection}

For this subtask, the participants are required to segment a set of untrimmed videos with the aim to retrieve strokes whatever the stroke class.
For this purpose, two annotated sets are provided: a training set and a validation  set, with respectively 16 and 6 untrimmed videos. A non-annotated test set consisting of 6 untrimmed videos has to be temporally segmented. The videos are not shared across the training, the validation, and test sets; however, the same player may appear in the different sets.
\par
Specifically, the participants have to fill the provided test set XML files with the stroke temporal boundaries (frame index of the videos).  All submissions will be evaluated in terms of mean Average Precision (mAP) and temporal Intersection over Union (IoU). Both are usually used for image segmentation but are adapted for this task:
\begin{itemize}
\item \textbf{mAP:} each stroke represents an object to be detected temporally. Detection is considered \texttt{True} when the temporal IoU between prediction and ground truth is above an IoU threshold. $20$ thresholds from $0.5$ to $0.95$ with a step of $0.05$ are considered, similarly to the COCO challenge~\cite{CocoChallenge2014}. This metric will be used for the final ranking of participants.
\item \textbf{IoU:} the frame-wise overlap between the ground truth and the predicted strokes across all the videos.
\end{itemize}
\par
For last year's edition, only two participants submitted runs for this difficult subtask~\cite{DBLP:conf/mediaeval/ZahraM21,DBLP:conf/mediaeval/JTBSN21}. They did not improve the baseline result in terms of mAP but \cite{DBLP:conf/mediaeval/JTBSN21} reached an IoU of 0.247 against 0.144 for the baseline, using YOLOv5 model.

\section{Dataset description}
\label{sec:dataset}

\begin{figure}
    \includegraphics[width=.32\linewidth]{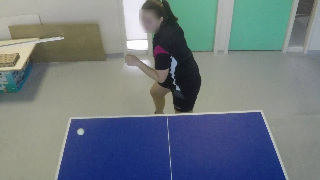}
    \includegraphics[width=.32\linewidth]{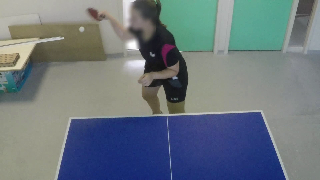}
    \includegraphics[width=.32\linewidth]{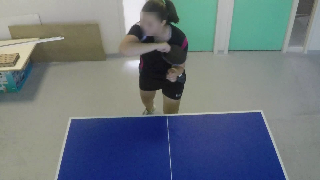}
    \caption{Key frames of a same stroke from \texttt{TTStroke-21}}
    \label{fig:dataset}
\end{figure}

The dataset was recorded at the Sports Faculty of the University of Bordeaux. It is constituted of player-centred videos without markers or sensors, recorded in natural conditions using GoPro cameras (see \autoref{fig:dataset}). Professional table tennis teachers designed a dedicated taxonomy to describe all the possible strokes. The dataset includes 20 table tennis stroke classes: 8 services, 6 offensive strokes, and 6 defensive strokes. The strokes may also be divided in two super-classes: \texttt{Forehand} and \texttt{Backhand}. The dataset was annotated by professional players using a crowd-sourced annotation platform. Non-stroke samples are inferred from the stroke annotations.
\par
In order to be able to share the dataset, we blurred the faces of the players for each original video frame using OpenCV deep learning face detector, based on the Single Shot Detector (SSD) framework with a ResNet base network. A tracking method has been implemented to decrease the false positive rate. The detected faces are blurred, and the video is re-encoded in MPEG-4.
\par
Compared with last year's edition, the classification dataset is enriched this year with new and more diverse video samples. The source videos were trimmed, sorted in class folders and distributed among train, validation and test sets. A total of 1~155 trimmed videos, representing more than 210~000 frames, are considered for this subtask. For the detection subtask, 100 minutes of table tennis games across 28 videos recorded at 120 frames per second and distributed in train, validation and test sets are considered. It represents more than $718$~$000$~frames. The resolution of the video for both subtasks is $1920\times1080$ representing in total 46.1~GB of disk space. The validation set is provided for each subtask for better comparison across participants. This set may be used for training when submitting the test set's results.

\section{Specific terms of use}
\label{sec:conditions}

Although faces are automatically blurred to preserve anonymity, some faces are misdetected, and thus some players remain identifiable. In order to respect the personal data of the players, this dataset is subject to a usage agreement, referred to as {\it Special Conditions}.
\par
These {\it Special Conditions} apply to the use of videos in the scope of the MediaEval workshop task ``Sport Task: Fine Grained Action Detection and Classification of Table Tennis Strokes from Videos.''. They correspond to the specific usage agreement referred to in the {\it Usage agreement for the MediaEval 2022 Research Collections}, signed between the user and the University of Delft. The complete acceptance of these {\it Special Conditions} is a mandatory prerequisite for the provision of the videos as part of the MediaEval 2022 evaluation campaign. A complete reading of these conditions is necessary and requires the user, for example, to obscure the faces (blurring, black banner) in the video before use in any publication and to destroy the data by January 30th, 2023.

\section{Discussions}
\label{sec:discussion}
As last year, the MediaEval Sport task offers two subtasks: i)~Classification and ii)~Detection of strokes. Classification methods performance has increased since the launch of the task. We hope to have more participation in the detection subtask in order to also improve in the domain of moment of interest in sports. The participants are encouraged to share their difficulties and their results even if they seem not sufficiently good. All the investigations, even when not successful, may inspire future methods.

\subsection*{Acknowledgment}
The authors would like to thank the players, coaches, and annotators who contributed to \texttt{TTStroke-21}.

\def\bibfont{\footnotesize} 
\bibliography{references} 

\begin{thebibliography}{25}
\expandafter\ifx\csname natexlab\endcsname\relax\def\natexlab#1{#1}\fi
\providecommand{\url}[1]{\texttt{#1}}
\providecommand{\href}[2]{#2}
\providecommand{\path}[1]{#1}
\providecommand{\DOIprefix}{doi:}
\providecommand{\ArXivprefix}{arXiv:}
\providecommand{\URLprefix}{URL: }
\providecommand{\Pubmedprefix}{pmid:}
\providecommand{\doi}[1]{\href{http://dx.doi.org/#1}{\path{#1}}}
\providecommand{\Pubmed}[1]{\href{pmid:#1}{\path{#1}}}
\providecommand{\bibinfo}[2]{#2}
\ifx\xfnm\relax \def\xfnm[#1]{\unskip,\space#1}\fi
\bibitem[{Martin et~al.(2021)Martin, Benois{-}Pineau, P{\'{e}}teri, and
  Morlier}]{PeICPR:2020}
\bibinfo{author}{P.~Martin}, \bibinfo{author}{J.~Benois{-}Pineau},
  \bibinfo{author}{R.~P{\'{e}}teri}, \bibinfo{author}{J.~Morlier},
\newblock \bibinfo{title}{3d attention mechanisms in twin spatio-temporal
  convolutional neural networks. application to action classification in videos
  of table tennis games.},
\newblock in: \bibinfo{booktitle}{{ICPR}}, \bibinfo{publisher}{{IEEE} Computer
  Society}, \bibinfo{year}{2021}.
\bibitem[{Martin(2020)}]{PeThesis2020}
\bibinfo{author}{P.~Martin}, \bibinfo{title}{Fine-Grained Action Detection and
  Classification from Videos with Spatio-Temporal Convolutional Neural
  Networks. Application to Table Tennis. (D{\'{e}}tection et classification
  fines d'actions {\`{a}} partir de vid{\'{e}}os par r{\'{e}}seaux de neurones
  {\`{a}} convolutions spatio-temporelles. Application au tennis de table)},
  Ph.D. thesis, University of La Rochelle, France, \bibinfo{year}{2020}.
  \URLprefix \url{https://tel.archives-ouvertes.fr/tel-03128769}.
\bibitem[{Martin et~al.(2021)Martin, Benois{-}Pineau, P{\'{e}}teri, Zemmari,
  and Morlier}]{PeChapSpringer:2021}
\bibinfo{author}{P.~Martin}, \bibinfo{author}{J.~Benois{-}Pineau},
  \bibinfo{author}{R.~P{\'{e}}teri}, \bibinfo{author}{A.~Zemmari},
  \bibinfo{author}{J.~Morlier}, \bibinfo{title}{3D Convolutional Networks for
  Action Recognition: Application to Sport Gesture Recognition},
  \bibinfo{publisher}{Springer International Publishing}, \bibinfo{year}{2021}.
\bibitem[{Soomro et~al.(2012)Soomro, Zamir, and Shah}]{Dataset:UCF101:2012}
\bibinfo{author}{K.~Soomro}, \bibinfo{author}{A.~R. Zamir},
  \bibinfo{author}{M.~Shah},
\newblock \bibinfo{title}{{UCF101:} {A} dataset of 101 human actions classes
  from videos in the wild},
\newblock \bibinfo{journal}{CoRR} \bibinfo{volume}{abs/1212.0402}
  (\bibinfo{year}{2012}).
\bibitem[{Kuehne et~al.(2011)Kuehne, Jhuang, Garrote, Poggio, and
  Serre}]{Dataset:HMDB:2011}
\bibinfo{author}{H.~Kuehne}, \bibinfo{author}{H.~Jhuang},
  \bibinfo{author}{E.~Garrote}, \bibinfo{author}{T.~A. Poggio},
  \bibinfo{author}{T.~Serre},
\newblock \bibinfo{title}{{HMDB:} {A} large video database for human motion
  recognition},
\newblock in: \bibinfo{booktitle}{{ICCV}}, \bibinfo{publisher}{{IEEE} Computer
  Society}, \bibinfo{year}{2011}, pp. \bibinfo{pages}{2556--2563}.
\bibitem[{Jhuang et~al.(2013)Jhuang, Gall, Zuffi, Schmid, and
  Black}]{Dataset:JHMDB:2013}
\bibinfo{author}{H.~Jhuang}, \bibinfo{author}{J.~Gall},
  \bibinfo{author}{S.~Zuffi}, \bibinfo{author}{C.~Schmid},
  \bibinfo{author}{M.~J. Black},
\newblock \bibinfo{title}{Towards understanding action recognition},
\newblock in: \bibinfo{booktitle}{{ICCV}}, \bibinfo{publisher}{{IEEE} Computer
  Society}, \bibinfo{year}{2013}, pp. \bibinfo{pages}{3192--3199}.
\bibitem[{Gu et~al.(2018)Gu, Sun, Ross, Vondrick, Pantofaru, Li,
  Vijayanarasimhan, Toderici, Ricco, Sukthankar, Schmid, and
  Malik}]{Dataset:AVA:2018}
\bibinfo{author}{C.~Gu}, \bibinfo{author}{C.~Sun}, \bibinfo{author}{D.~A.
  Ross}, \bibinfo{author}{C.~Vondrick}, \bibinfo{author}{C.~Pantofaru},
  \bibinfo{author}{Y.~Li}, \bibinfo{author}{S.~Vijayanarasimhan},
  \bibinfo{author}{G.~Toderici}, \bibinfo{author}{S.~Ricco},
  \bibinfo{author}{R.~Sukthankar}, \bibinfo{author}{C.~Schmid},
  \bibinfo{author}{J.~Malik},
\newblock \bibinfo{title}{{AVA:} {A} video dataset of spatio-temporally
  localized atomic visual actions}  (\bibinfo{year}{2018})
  \bibinfo{pages}{6047--6056}.
\bibitem[{Kay et~al.(2017)Kay, Carreira, Simonyan, Zhang, Hillier,
  Vijayanarasimhan, Viola, Green, Back, Natsev, Suleyman, and
  Zisserman}]{Dataset:Kinetics:2017}
\bibinfo{author}{W.~Kay}, \bibinfo{author}{J.~Carreira},
  \bibinfo{author}{K.~Simonyan}, \bibinfo{author}{B.~Zhang},
  \bibinfo{author}{C.~Hillier}, \bibinfo{author}{S.~Vijayanarasimhan},
  \bibinfo{author}{F.~Viola}, \bibinfo{author}{T.~Green},
  \bibinfo{author}{T.~Back}, \bibinfo{author}{P.~Natsev},
  \bibinfo{author}{M.~Suleyman}, \bibinfo{author}{A.~Zisserman},
\newblock \bibinfo{title}{The kinetics human action video dataset},
\newblock \bibinfo{journal}{CoRR} \bibinfo{volume}{abs/1705.06950}
  (\bibinfo{year}{2017}).
\bibitem[{Smaira et~al.(2020)Smaira, Carreira, Noland, Clancy, Wu, and
  Zisserman}]{Dataset:Kinetics700:2020}
\bibinfo{author}{L.~Smaira}, \bibinfo{author}{J.~Carreira},
  \bibinfo{author}{E.~Noland}, \bibinfo{author}{E.~Clancy},
  \bibinfo{author}{A.~Wu}, \bibinfo{author}{A.~Zisserman},
\newblock \bibinfo{title}{A short note on the kinetics-700-2020 human action
  dataset},
\newblock \bibinfo{journal}{CoRR} \bibinfo{volume}{abs/2010.10864}
  (\bibinfo{year}{2020}).
\bibitem[{Li et~al.(2020)Li, Thotakuri, Ross, Carreira, Vostrikov, and
  Zisserman}]{Dataset:AVA_Kinetics:2020}
\bibinfo{author}{A.~Li}, \bibinfo{author}{M.~Thotakuri}, \bibinfo{author}{D.~A.
  Ross}, \bibinfo{author}{J.~Carreira}, \bibinfo{author}{A.~Vostrikov},
  \bibinfo{author}{A.~Zisserman},
\newblock \bibinfo{title}{The ava-kinetics localized human actions video
  dataset},
\newblock \bibinfo{journal}{CoRR} \bibinfo{volume}{abs/2005.00214}
  (\bibinfo{year}{2020}).
\bibitem[{Shao et~al.(2020)Shao, Zhao, Dai, and Lin}]{Dataset:Gym:2020}
\bibinfo{author}{D.~Shao}, \bibinfo{author}{Y.~Zhao}, \bibinfo{author}{B.~Dai},
  \bibinfo{author}{D.~Lin},
\newblock \bibinfo{title}{Finegym: {A} hierarchical video dataset for
  fine-grained action understanding},
\newblock in: \bibinfo{booktitle}{{CVPR}}, \bibinfo{publisher}{{IEEE}},
  \bibinfo{year}{2020}, pp. \bibinfo{pages}{2613--2622}.
\bibitem[{Martin et~al.(2020)Martin, Benois{-}Pineau, P{\'{e}}teri, and
  Morlier}]{PeMTAP:2020}
\bibinfo{author}{P.~Martin}, \bibinfo{author}{J.~Benois{-}Pineau},
  \bibinfo{author}{R.~P{\'{e}}teri}, \bibinfo{author}{J.~Morlier},
\newblock \bibinfo{title}{Fine grained sport action recognition with twin
  spatio-temporal convolutional neural networks},
\newblock \bibinfo{journal}{Multim. Tools Appl.} \bibinfo{volume}{79}
  (\bibinfo{year}{2020}) \bibinfo{pages}{20429--20447}.
\bibitem[{Liu and Hu(2019)}]{LiuH:2019}
\bibinfo{author}{Z.~Liu}, \bibinfo{author}{H.~Hu},
\newblock \bibinfo{title}{Spatiotemporal relation networks for video action
  recognition},
\newblock \bibinfo{journal}{{IEEE} Access} \bibinfo{volume}{7}
  (\bibinfo{year}{2019}) \bibinfo{pages}{14969--14976}.
\bibitem[{Carreira and Zisserman(2017)}]{NN:I3DCarreira:2017}
\bibinfo{author}{J.~Carreira}, \bibinfo{author}{A.~Zisserman},
\newblock \bibinfo{title}{Quo vadis, action recognition? {A} new model and the
  kinetics dataset},
\newblock in: \bibinfo{booktitle}{{CVPR}}, \bibinfo{publisher}{{IEEE} Computer
  Society}, \bibinfo{year}{2017}, pp. \bibinfo{pages}{4724--4733}.
\bibitem[{Varol et~al.(2018)Varol, Laptev, and Schmid}]{NN:Laptev:2018}
\bibinfo{author}{G.~Varol}, \bibinfo{author}{I.~Laptev},
  \bibinfo{author}{C.~Schmid},
\newblock \bibinfo{title}{Long-term temporal convolutions for action
  recognition},
\newblock \bibinfo{journal}{{IEEE} Trans. Pattern Anal. Mach. Intell.}
  \bibinfo{volume}{40} (\bibinfo{year}{2018}) \bibinfo{pages}{1510--1517}.
\bibitem[{Martin et~al.(2019)Martin, Benois{-}Pineau, P{\'{e}}teri, and
  Morlier}]{PeICIP:2019}
\bibinfo{author}{P.~Martin}, \bibinfo{author}{J.~Benois{-}Pineau},
  \bibinfo{author}{R.~P{\'{e}}teri}, \bibinfo{author}{J.~Morlier},
\newblock \bibinfo{title}{Optimal choice of motion estimation methods for
  fine-grained action classification with 3d convolutional networks},
\newblock in: \bibinfo{booktitle}{{ICIP}}, \bibinfo{publisher}{{IEEE}},
  \bibinfo{year}{2019}, pp. \bibinfo{pages}{554--558}.
\bibitem[{Martin et~al.(2018)Martin, Benois{-}Pineau, P{\'{e}}teri, and
  Morlier}]{PeCBMI:2018}
\bibinfo{author}{P.~Martin}, \bibinfo{author}{J.~Benois{-}Pineau},
  \bibinfo{author}{R.~P{\'{e}}teri}, \bibinfo{author}{J.~Morlier},
\newblock \bibinfo{title}{Sport action recognition with siamese spatio-temporal
  cnns: Application to table tennis},
\newblock in: \bibinfo{booktitle}{{CBMI}}, \bibinfo{publisher}{{IEEE}},
  \bibinfo{year}{2018}, pp. \bibinfo{pages}{1--6}.
\bibitem[{Martin et~al.(2021)Martin, Calandre, Mansencal, Benois{-}Pineau,
  P{\'{e}}teri, Mascarilla, and Morlier}]{DBLP:conf/mediaeval/MartinCMBPMM21}
\bibinfo{author}{P.~Martin}, \bibinfo{author}{J.~Calandre},
  \bibinfo{author}{B.~Mansencal}, \bibinfo{author}{J.~Benois{-}Pineau},
  \bibinfo{author}{R.~P{\'{e}}teri}, \bibinfo{author}{L.~Mascarilla},
  \bibinfo{author}{J.~Morlier},
\newblock \bibinfo{title}{Sports video: Fine-grained action detection and
  classification of table tennis strokes from videos for mediaeval 2021},
\newblock in:  \cite{DBLP:conf/mediaeval/2021}, \bibinfo{year}{2021}.
  \URLprefix \url{http://ceur-ws.org/Vol-3181/paper3.pdf}.
\bibitem[{Martin(2022)}]{mediaeval/Martin/2022/baseline}
\bibinfo{author}{P.~Martin},
\newblock \bibinfo{title}{Baseline method for the sport task of mediaeval 2022
  benchmark with 3d cnns using attention mechanism},
\newblock in: \bibinfo{booktitle}{MediaEval}, {CEUR} Workshop Proceedings,
  \bibinfo{publisher}{CEUR-WS.org}, \bibinfo{year}{2022}.
\bibitem[{Qian et~al.(2021)Qian, Yu, Liu, and
  Hauptmann}]{DBLP:conf/mediaeval/QianYL021}
\bibinfo{author}{Y.~Qian}, \bibinfo{author}{L.~Yu}, \bibinfo{author}{W.~Liu},
  \bibinfo{author}{A.~Hauptmann},
\newblock \bibinfo{title}{Learning unbiased transformer for long-tail sports
  action classification},
\newblock in:  \cite{DBLP:conf/mediaeval/2021}, \bibinfo{year}{2021}.
  \URLprefix \url{http://ceur-ws.org/Vol-3181/paper52.pdf}.
\bibitem[{Martin(2021)}]{DBLP:conf/mediaeval/Martin21}
\bibinfo{author}{P.~Martin},
\newblock \bibinfo{title}{Spatio-temporal {CNN} baseline method for the sports
  video task of mediaeval 2021 benchmark},
\newblock in:  \cite{DBLP:conf/mediaeval/2021}, \bibinfo{year}{2021}.
  \URLprefix \url{http://ceur-ws.org/Vol-3181/paper13.pdf}.
\bibitem[{Lin et~al.(2014)Lin, Maire, Belongie, Hays, Perona, Ramanan,
  Doll{\'{a}}r, and Zitnick}]{CocoChallenge2014}
\bibinfo{author}{T.~Lin}, \bibinfo{author}{M.~Maire}, \bibinfo{author}{S.~J.
  Belongie}, \bibinfo{author}{J.~Hays}, \bibinfo{author}{P.~Perona},
  \bibinfo{author}{D.~Ramanan}, \bibinfo{author}{P.~Doll{\'{a}}r},
  \bibinfo{author}{C.~L. Zitnick},
\newblock \bibinfo{title}{Microsoft {COCO:} common objects in context},
\newblock in: \bibinfo{editor}{D.~J. Fleet}, \bibinfo{editor}{T.~Pajdla},
  \bibinfo{editor}{B.~Schiele}, \bibinfo{editor}{T.~Tuytelaars} (Eds.),
  \bibinfo{booktitle}{Computer Vision - {ECCV} 2014 - 13th European Conference,
  Zurich, Switzerland, September 6-12, 2014, Proceedings, Part {V}}, volume
  \bibinfo{volume}{8693} of \textit{\bibinfo{series}{Lecture Notes in Computer
  Science}}, \bibinfo{publisher}{Springer}, \bibinfo{year}{2014}, pp.
  \bibinfo{pages}{740--755}.
\bibitem[{Zahra and Martin(2021)}]{DBLP:conf/mediaeval/ZahraM21}
\bibinfo{author}{A.~Zahra}, \bibinfo{author}{P.~Martin},
\newblock \bibinfo{title}{Two stream network for stroke detection in table
  tennis},
\newblock in:  \cite{DBLP:conf/mediaeval/2021}, \bibinfo{year}{2021}.
  \URLprefix \url{http://ceur-ws.org/Vol-3181/paper55.pdf}.
\bibitem[{J et~al.(2021)J, T, B, S, and N}]{DBLP:conf/mediaeval/JTBSN21}
\bibinfo{author}{B.~J}, \bibinfo{author}{M.~T. T}, \bibinfo{author}{B.~B},
  \bibinfo{author}{J.~S}, \bibinfo{author}{L.~N. N},
\newblock \bibinfo{title}{{YOLOV5} for stroke detection and classification in
  table tennis},
\newblock in:  \cite{DBLP:conf/mediaeval/2021}, \bibinfo{year}{2021}.
  \URLprefix \url{http://ceur-ws.org/Vol-3181/paper38.pdf}.
\bibitem[{Hicks et~al.(2022)Hicks, Pogorelov, Lommatzsch, de~Herrera, Martin,
  Hassan, Porter, Kasem, Andreadis, Lux, Oca{\~{n}}a, Liu, and
  Larson}]{DBLP:conf/mediaeval/2021}
\bibinfo{editor}{S.~Hicks}, \bibinfo{editor}{K.~Pogorelov},
  \bibinfo{editor}{A.~Lommatzsch}, \bibinfo{editor}{A.~G.~S. de~Herrera},
  \bibinfo{editor}{P.~Martin}, \bibinfo{editor}{S.~Z. Hassan},
  \bibinfo{editor}{A.~Porter}, \bibinfo{editor}{A.~Kasem},
  \bibinfo{editor}{S.~Andreadis}, \bibinfo{editor}{M.~Lux},
  \bibinfo{editor}{M.~G. Oca{\~{n}}a}, \bibinfo{editor}{A.~Liu},
  \bibinfo{editor}{M.~Larson} (Eds.), \bibinfo{title}{Working Notes Proceedings
  of the MediaEval 2021 Workshop, Online, 13-15 December 2021}, volume
  \bibinfo{volume}{3181} of \textit{\bibinfo{series}{{CEUR} Workshop
  Proceedings}}, \bibinfo{publisher}{CEUR-WS.org}, \bibinfo{year}{2022}.
  \URLprefix \url{http://ceur-ws.org/Vol-3181}.

\end{thebibliography}

\end{document}